\documentclass[conference]{IEEEtran}
\IEEEoverridecommandlockouts

\usepackage{cite}
\usepackage{amsmath,amssymb,amsfonts}
\usepackage{graphicx}
\usepackage{textcomp}
\usepackage{xcolor}
\usepackage{booktabs}
\usepackage{multirow}
\usepackage{url}
\usepackage{microtype}
\usepackage{color,soul}
\usepackage{enumitem}
\usepackage{algorithm}
\usepackage{algpseudocode}
\usepackage{fancyhdr}

\setlist{noitemsep, topsep=2pt}

\def\BibTeX{{\rm B\kern-.05em{\sc i\kern-.025em b}\kern-.08em
    T\kern-.1667em\lower.7ex\hbox{E}\kern-.125emX}}

\newcommand\freefootnote[1]{%
  \let\svthefootnote\thefootnote%
  \let\thefootnote\relax%
  \footnotetext{#1}%
  \let\thefootnote\svthefootnote%
}

\usepackage{titlesec}
\titlespacing*{\section}{0pt}{4pt}{4pt}
\titlespacing*{\subsection}{0pt}{5pt}{4pt}

\begin{document}

\title{Design Space Exploration of Reinforcement Learning Configurations for Post-Legalization Placement}

\author{
  \IEEEauthorblockN{
    Foteini~Oikonomou\textsuperscript{*},~Maria~Pantazi-Kypraiou\textsuperscript{*},~George~Goudroumanis\textsuperscript{*},~Pavlos~Stoikos\textsuperscript{*}, \\ ~Athanasios~Tziouvaras\textsuperscript{*}, ~Christos~Anagnostopoulos\textsuperscript{†}, ~Georgios~Karakonstantis\textsuperscript{*}, ~and~George~Floros\textsuperscript{‡}
  }
  \IEEEauthorblockA{
    \textsuperscript{*}Department of Electrical and Computer Engineering,
    University of Thessaly, Volos, Greece \\
    emails: \{fotoikon, mpantazi-, ggeorgios-r, pastoikos, attziouv, gkarakon\}@e-ce.uth.gr
  }
  \IEEEauthorblockA{
    \textsuperscript{†}School of Computing Science,
    University of Glasgow, UK, email: christos.anagnostopoulos@glasgow.ac.uk
  }
  \IEEEauthorblockA{
    \textsuperscript{‡}Department of Electronic and Electrical Engineering,
    Trinity College Dublin, Ireland, email: florosg@tcd.ie
  }
}

\maketitle
\begin{abstract}
Graph Neural Networks (GNNs) and Reinforcement Learning (RL) have recently been explored for VLSI placement optimization, but their effectiveness for post-legalization placement refinement remains unclear. This paper presents a systematic design space exploration of a GNN-Proximal Policy Optimization (GNN-PPO) framework across three ISPD2015 benchmarks. We evaluate 18 configurations covering reward functions, GNN architectures, action spaces, candidate-set sizes, and training strategies. We analyse the results we obtain to identify pitfalls and to assess the prospects and opportunities of the RL methods for post-legalization placement. On top of that we propose a design classification approach to guide the design decisions of RL models.
\end{abstract}


\section{Introduction}

Post-legalization detailed placement refinement seeks to reduce half-perimeter
wirelength (HPWL) after global placement and legalization, without violating site constraints. Classical methods such as perturbation moves, local search, and HPWL-greedy swaps have been refined over decades and remain highly competitive \cite{openroad2021,ispd2015contest}. Analytical placers such as DREAMPlace \cite{lin2019dreamplace} and RePlAce \cite{cheng2019replace} achieve near-optimal results via differentiable HPWL optimization. Reinforcement learning (RL) approaches, popularized by macro placement \cite{alphachip2021}, emerge as an alternative approach. However, RL-based standard-cell methods
\cite{agnesina2022parameter,ren2021standard} remain underexplored and operate at smaller scale or under simplified assumptions, such as discretized grids.

\freefootnote{*This work was funded under the COIN-3D project, which has received funding from the European Union's Horizon Europe research and innovation program under grant agreement No. 101159667.}

The direct application of RL to standard-cell refinement is therefore poorly
characterized. It is unclear which GNN architecture, reward formulation, or action space is most effective, and whether the RL can scale to large designs. This paper addresses that gap with a controlled, multi-dimensional design space exploration across three ISPD2015 benchmarks. It reports on the obtained results and benchmarks its findings against OpenROAD \cite{openroad2021}.


The paper makes four main contributions:

\begin{itemize}
    \item We conduct a systematic design space exploration of an RL method that
    combines a GNN with PPO for post-legalization placement refinement, evaluating
    18 configurations across several design dimensions and over both regular
    and irregular netlists.
    \item We uncover pitfalls and design considerations that, if left unchecked,
    may impede the effectiveness of RL, and we show they are tightly linked to
    whether the netlist is regular or irregular.
    \item We identify the prospects and opportunities of RL and show that graph
    partitioning overcomes some of its limitations where the full-graph reward is
    noisy or sparse, helping substantially on an irregular design and modestly on
    a regular one.
    \item We propose a design classification method that provides practical
    guidance for choosing the most suitable training strategy.
\end{itemize}

\section{Design space exploration framework}
\label{sec:method}
In this section we elaborate on the different methods, models and parameters we use in order to perform the envisioned design space exploration.

\subsection{Problem Formulation}
In reinforcement learning, an agent is leveraged that observes an environment, takes some actions and then, receives a reward that resembles the utility of selected actions. In our case, given a legalized placement $NET$ of $N$ standard cells, the agent selects a sequence of single-cell moves to minimize total HPWL. Thus, for each time step $t$, the RL follows an iterative process which consists the following operations:
\begin{enumerate}
  \item \textbf{Observation:} $NET$ is fed to a GNN, which outputs the representation of the RL environment.
  \item \textbf{Refinement:} Each cell is assigned a score by the GNN. The top-$K$ cells (in terms of that score) formulate candidate set, upon which the agent will choose an action.
  \item \textbf{Action selection:} The policy $\pi_\theta$ selects one cell $c$ and one action $a$.
  \item \textbf{Action implementation:} Cell $c$ moves according to action $a$. We allow temporary overlaps during moves, which we then fix with a final legalization step.

  \item \textbf{Reward generation:} A reward $r_t$ is computed and the agent updates its policy via $\pi_\theta$ the PPO technique.
\end{enumerate}

\subsection{Graph Construction}
The netlist is represented as a heterogeneous graph $\mathcal{G} = (V, E)$ where
nodes $v_i \in V$ are placeable cells with features $(x_i, y_i, w_i, h_i)$ and
edges $(v_i, v_j) \in E$ for every two-pin net. High-fanout nets ($>$20 pins), power pins and clock nets are excluded from edge construction to avoid dominating the GNN. 

\subsection{GNN Architecture}
We evaluate the effectiveness of three GNN architectures:
\begin{itemize}
\item \textbf{GAT-2L:} A standard graph attention model with $2$ layers.
\item \textbf{GCN:} A spectral convolution architecture implemented in  \cite{kipf2017gcn}.
\item \textbf{GraphSAGE:} The  GraphSAGE architecture, implemented in  \cite{hamilton2017sage}.
\end{itemize}

\subsection{Action spaces}
We evaluate three action spaces:
\begin{itemize}
\item \textbf{1-site move:} Move the cell to one adjacent placement site; $a \in \{$N, S, E, W, no\_move$\}$.
\item \textbf{3-site move:} Move the cell to a position, which is located $3$ placement sites away; $a \in \{$NNN, SSS, EEE, WWW, no\_move$\}$.
\item \textbf{Multi-site move:} Move the cell to a position, which is located up to $3$ placements sites away; $a \in \{$N, NN, NNN, S, SS, SSS, E, EE, EEE, W, WW, WWW, no\_move$\}$.
\end{itemize}


\subsection{Reward function}
In this work, we analyze the effect of different reward structures in order to study their effects, as follows. The first reward structure, named \textbf{Local} is derived as follows:

\begin{equation}
Local = \textstyle\sum_{n \in \mathcal{N}_t} \big( \text{HPWL}_n(P_{t-1}) - \text{HPWL}_n(P_t) \big)
\label{eq:reward}
\end{equation}

where $\mathcal{N}_t$ is the set of the nets incident to the moved cell $c$, during the step $t$.

The second reward structure, named \textbf{Normalized} is derived as follows:
\begin{equation}
Normalized = \frac{Local}{d}
\label{eq:reward_normalized}
\end{equation}

where $d$ is the total perimeter of the die area.

\subsection{Graph partitioning}
\label{sec:subregion_method}
For large or medium designs where the full-graph reward signal is too sparse,
we introduce a \emph{subregion training} technique: For each RL iteration, a random bounding box is generated that covers a fraction (denoted as $f \in [0,1]$) of the die area and then, the agent operates on this sampled graph partition. Evidently, when $f \to 0$ the agent has access to a smaller graph segment; whereas when $f \to 1$ the agent investigates larger graph segments in a single run. In both cases, any nets that remain outside of the subgraph are treated as virtual external nodes and contribute normally to the reward generation, but they are excluded from the action space.


\section{Experimental setup}
For the design space exploration we leverage the ISPD2015 designs \cite{ispd2015contest}, as depicted in Table~\ref{tab:designs}. Benchmarks exhibit
distinct structural properties that directly affect GNN performance, as discussed in section \ref{sec:scalability}. 
Starting from post-legalized layouts, refinement moves may cause overlaps, which are resolved via a final legalization pass. OpenROAD \cite{openroad2021} serves as our baseline.

\begin{table}[t]
\caption{The ISPD2015 benchmarks used to perform the design space exploration.}
\label{tab:designs}
\centering
\scriptsize
\setlength{\tabcolsep}{13pt}
\begin{tabular}{lcc}
\toprule
Design & $N$ (cells) & Baseline HPWL ($\mu$m)  \\
\midrule
\texttt{mgc\_fft\_1}         &  32K  & 260,172  \\
\texttt{mgc\_pci\_bridge32\_a} & ${\sim}$30K  & 548,355 B \\
\texttt{mgc\_matrix\_mult\_1} & 155K  & 1,518,463  \\
\bottomrule
\end{tabular}
\end{table}

We run the RL training process, for each different RL configuration for $400$ episodes ($800$ when we use our graph partitioning technique). For the evaluation process we perform inference, using the trained RL agent over the corresponding benchmark. To evaluate the performance of each RL configuration we use the following metrics:

\begin{itemize}
    \item \textbf{Training peak:} The episode during training where the agent achieves the best HPWL decrease (\textit{{Train$\Delta$\%}}).
    \item \textbf{Inference score:} The decrease of HPWL (\textit{{Eval$\Delta$\%}}) that is achieved during inference, using the best performing RL checkpoint.
\end{itemize}

\section{Experimental evaluation}
\label{sec:dse}


Table~\ref{tab:experiments} summarizes the conducted experiments along with the corresponding \textit{{Train$\Delta$\%}} and \textit{{Eval$\Delta$\%}}. For reference, OpenROAD achieved an HPWL reduction of $6.4\%$  for \texttt {mgc\_fft\_1}, $4.3\%$ for \texttt {mgc\_pci\_bridge32\_a}, and $4.6\%$ for \texttt {mgc\_matrix\_mult\_1}. The results we obtain expose some key insights of the RL underlying parameters.

\begin{table}[h]
\caption{Design space exploration configurations and results, grouped per
benchmark. }
\label{tab:experiments}
\centering
\scriptsize
\setlength{\tabcolsep}{3pt}
\begin{tabular}{cccccccc}
\hline
Config. & Arch. & Action & Reward & $K$ & Partit. & Train $\Delta$\% & Eval $\Delta$\% \\
\hline

\multicolumn{8}{l}{\texttt{mgc\_fft\_1}} \\
C1   & GAT-2L & 1-site     & Local           & 5K  & No & $0.44$ & $0.20$  \\
C2   & GAT-2L & 1-site     & Normalized      & 5K  & No & $0.41$ & $0.20$ \\
C3   & GCN    & 1-site     & Local           & 5K  & No & $0.46$ & $0.21$ \\
C4   & SAGE   & 1-site     & Local           & 5K  & No & $\mathbf{0.53}$ & $\mathbf{0.20}$ \\
C5   & GAT-2L & 3-site     & Local           & 5K  & No & $0.26$ & $0.22$ \\
C7   & SAGE   & 1-site     & Local           & 5K  & $f{=}0.15$ & $0.01$ & $0.00$ \\
\midrule
\multicolumn{8}{l}{\texttt{mgc\_pci\_bridge32\_a}} \\
C8  & GAT-2L & 1-site     & Local           & 5K  & No & $0.36$ & $0.15$ \\
C9  & SAGE   & 1-site     & Local           & 5K  & No & $0.00$ & $0.00$ \\
C10 & GAT-2L & 1-site     & Local           & 5K  & $f{=}0.15$ & $2.99$ & $0.15$ \\
C11 & SAGE   & 1-site     & Local           & 5K  & $f{=}0.15$ & $\mathbf{5.85}$ & $\mathbf{0.43}$ \\
\midrule
\multicolumn{8}{l}{\texttt{mgc\_matrix\_mult\_1}} \\
C12  & GAT-2L & 1-site     & Local           & 5K  & No & $0.01$ & $0.01$ \\
C13  & GAT-2L & 1-site     & Local           & 20K & No & $0.00$ & $0.01$ \\
C14  & SAGE   & 1-site     & Local           & 5K  & No & $0.01$ & $0.01$ \\
C15  & GAT-2L & 1-site     & Scaled          & 5K  & No & $0.00$ & $0.01$ \\
C16  & GAT-2L & 1-site     & Per-Net         & 5K  & No & $0.00$ & $0.01$ \\
C17  & SAGE   & 1-site     & Local           & 5K  & $f{=}0.15$ & $0.03$ & $0.01$ \\
C18  & SAGE   & 1-site     & Local           & 5K  & $f{=}0.08$ & $\mathbf{0.04}$ & $\mathbf{0.03}$ \\
\bottomrule
\end{tabular}
\smallskip
\end{table}

\textbf{GNN architecture.} GraphSAGE achieves some of the best training performance, but exhibits policy collapse in cases where the model is over-trained. In such cases, the reward degrades after and the model's best checkpoint is not realized in the last training round. On the other hand, GCN converges more slowly yet remains stable throughout the training process. In terms of inference, all architectures produce comparable results and thus, the training advantage of GraphSAGE is not fully transferred to inference.


\textbf{Action space.} We also tried a \textbf{Swap} option, which exchanges the selected cell with its nearest neighbor, however it is computationally expensive and difficult to converge. Among feasible variants, the \textbf{3-site move} marginally outperforms \textbf{1-site move} during inference ($0.22$ vs $0.20$ Eval $\Delta$\% for the \texttt{mgc\_fft\_1} design), though demonstrates small gains.

\textbf{Reward function.} The \textbf{Local} and \textbf{Normalized} rewards achieve comparable results and do not seem to significantly affect the RL training or inference process.  A \textbf{Global} reward recomputing the total HPWL each move matches the exact objective but is ${\sim}100\times$ slower per episode, so we discarded it as intractable. We have also explored additional reward engineering approaches for the \texttt{mgc\_matrix\_mult\_1} benchmark in order to increase the RL performance. To this end, two additional rewards are explored, just for this benchmark. The first one, denoted as \textbf{Per-Net} is defined as follows:

\begin{equation}
\textit{Per-Net} = \frac{Local}{\frac{1}{|\mathcal{N}_t|}\sum_{n \in \mathcal{N}_t} \text{HPWL}_n(P_{t-1})}
\label{eq:pernet}
\end{equation}

\noindent that is, the \textbf{Local} reward normalized by the mean pre-move HPWL of the nets incident to the moved cell, making the reward magnitude scale-invariant across die sizes,
and the second one, denoted as \textbf{Scaled}, defined as: $Scaled = a * Local$, which is a essentially an amplified Local reward, by a factor of $a$. In our case, we set $a=6$. Both of these reward structures fail to deliver a better RL performance, thus highlighting that reward engineering cannot alone solve the RL training challenge.



\textbf{Graph partitioning.} Graph partitioning plays an important role on the overall RL performance, but its benefit is interconnected with the micro-architecture of each design and with the reason why full-graph training struggles. For example, the \texttt{mgc\_pci\_bridge32\_a} benchmark implements a bus circuit, where the bottleneck is a noisy reward signal. Restricting the agent to a small bounding box concentrates the local signal and stabilizes the learning process, and partitioning is therefore highly effective here. On the other hand, the \texttt{mgc\_matrix\_mult\_1} benchmark implements a circuit for matrix multiplication ($\mathbf{C}{=}\mathbf{A}{\times}\mathbf{B}$), designed via a regular partial-product array where every cell shares a similar $k$-hop neighborhood structure. As a result, a policy trained on a single partition can easily generalize to the rest of the design, since the cell topology is repeated in a periodic pattern, which makes even very small partitions effective and cheap to train. However, global placement has already optimized this regular layout, so little is left to improve and partitioning brings only a small gain. Partitioning also fails on \texttt{mgc\_fft\_1}, its butterfly topology relies on long-range connections, so cutting the design into small boxes severs these nets and destroys the local reward signal, causing the policy to collapse.



\textbf{Comparison with OpenROAD.} OpenROAD achieves $4.3\%$ to $6.4\%$ HPWL reduction, above the final inference results of our RL configurations. The reason is largely structural, since OpenROAD evaluates coordinated multi-cell moves against the exact global HPWL, whereas the RL refines one cell at a time from a local signal, so credit assignment (Section~\ref{sec:scalability}) is harder. The RL still shows promise, as on the irregular \texttt{mgc\_pci\_bridge32\_a} several training checkpoints exceed OpenROAD, so the main limitation is not solution quality but policy stability at inference. Reducing the search space through subregion training improves learning, and our netlist classification exploits this to flag designs that benefit from partitioned optimization.

\section{Pitfalls and RL design considerations}
\label{sec:scalability}
In this section we elaborate on the pitfalls of the RL design phase, which have the capacity to degrade the model performance, if left unaddressed by the designers.

\textbf{(1) Credit assignment bottleneck.} 
Given the fact that the agent examines only portion of the total cells per episode ($K/N$), it has a limited perception of the whole netlist. This phenomenon is apparent in \texttt{mgc\_matrix\_mult\_1}, where the agent covers $3.2\%$ of the total cells per episode, versus $15.5\%$ for \texttt{mgc\_fft\_1}, and $16.9\%$ for \texttt{mgc\_pci\_bridge32\_a }. In this case, the PPO fails to assign a policy to rest of the $96.8\%$ of cells in each episode and thus, the model does not converge. Evidently, higher coverage rates (as in \texttt{mgc\_pci\_bridge32\_a} and \texttt{mgc\_fft\_1}) solve this problem, but require a higher $K$ value. This translates to higher computational complexity and larger training times.

\textbf{(2) GNN over-smoothing.} The micro-architecture of \texttt{mgc\_matrix\_mult\_1}, and more specifically the fact that its cells share a similar neighborhood structure, can lead to over-smoothing. This phenomenon emerges when the GNN produces near-identical embeddings for all cells thus, preventing the policy from distinguishing beneficial moves from harmful ones. In contrast, the wire connections in \texttt{mgc\_fft\_1} connections create structurally unique multi-hop neighborhoods and result in diverse embeddings. Overall, in designs where several cells share the same neighborhood configuration, GNN seem to decrease the model quality.
 
\textbf{(3) Tightly placed cells.} The periodic grid structure of a matrix multiplier maps well to a regular silicon layout, which Global placement exploits efficiently. The few remaining misplacements are largely out of reach for single-cell local moves, so the RL finds little to improve even though coordinated multi-cell refiners still reduce HPWL on this design. On the other hand, \texttt{mgc\_fft\_1} and \texttt{mgc\_pci\_bridge32\_a} carry more local residual misplacements that single-cell RL moves can exploit.




\section{Prospects and opportunities of RL}
\label{sec:regime}
Results indicate that the RL can benefit from prior knowledge on each design's characteristics (e.g. micro-architectural details). This is apparent when the agent is trained using smaller graph partitions, instead of the whole circuit. Then a question arises on how to a extract such information from a given netlist, without requiring to train the RL agent for a large number of steps. One possible solution is to classify the netlist to a set of pre-defined classes, which indicate how the graph should be partitioned. In this works we perform a preliminary assessment of this approach and we devise three distinct classes as follows:
\begin{itemize}
    \item Class A: The graph should not be partitioned. 
    \item Class B: The graph should be partitioned, with $f = 0.15$.
    \item Class C: The graph should be partitioned, with $f = 0.08$.
\end{itemize}




Given a netlist \texttt{net}, we assign it a static \emph{partition score} $S_p$:
\begin{equation}
  S_p = \frac{\overline{\text{HPWL}}_{\text{net}}}{d} \times \frac{K}{N}
  \label{eq:score}
\end{equation}

\noindent where $\overline{\text{HPWL}}_{\text{net}}$ is the mean per-net HPWL (baseline HPWL divided by the number of nets), $d$ the die perimeter, $K$ the candidate-set size, and $N$ the number of cells. The first factor is the per-move reward magnitude and the second the fraction of cells acted on per episode.

A high $S_p$ alone does not guarantee a usable signal, since the reward can still be dominated by noise. We thus add a \emph{dynamic} metric $R_m$, the reward signal-to-noise ratio over a few short random-policy probes without training:
\begin{equation}
  R_m = \frac{|\mu_R|}{\sigma_R}
  \label{eq:snr}
\end{equation}
where $\mu_R$ and $\sigma_R$ are the mean and standard deviation of the reward for the episode. A low $R_m$ means noisy rewards and unreliable gradient updates. Algorithm \ref{alg:class} uses both metrics to assign each design to a class (A, B, or C) and its thresholds are calibrated on our three designs and are indicative.



\begin{algorithm}[t]
\caption{Netlist classification}\label{alg:class}
\begin{algorithmic}[1]
\Require netlist \texttt{net}; candidate-set size $K$; score threshold $\theta_s{=}5\times10^{-4}$; SNR threshold $\theta_m{=}2.0$; $P{=}5$ probes of $L{=}500$ steps
\Ensure class $\in \{$A, B, C$\}$ and partition fraction $f$
\State $N \gets$ number of cells in \texttt{net}; \, $d \gets$ die perimeter
\State $\overline{\text{HPWL}}_{\text{net}} \gets$ baseline HPWL of \texttt{net} $/$ number of nets
\State $S_p \gets (\overline{\text{HPWL}}_{\text{net}} / d) \times (K / N)$ \Comment{Eq.~\eqref{eq:score}}
\State $\mathcal{R} \gets \emptyset$
\For{$i \gets 1$ to $P$}
    \State $r_i \gets$ total reward of an $L$-step random-policy rollout
    \State $\mathcal{R} \gets \mathcal{R} \cup \{r_i\}$ \Comment{no training}
\EndFor
\State $R_m \gets |\text{mean}(\mathcal{R})| / \text{std}(\mathcal{R})$ \Comment{Eq.~\eqref{eq:snr}}
\If{$R_m < \theta_m$}
    \State class $\gets$ B; \, $f \gets 0.15$ \Comment{noisy reward}
\ElsIf{$S_p > \theta_s$}
    \State class $\gets$ A; \, $f \gets 1$ \Comment{strong, clean signal}
\Else
    \State class $\gets$ C; \, $f \gets 0.08$ \Comment{weak signal}
\EndIf
\State \Return (class, $f$)
\end{algorithmic}
\end{algorithm}



In order to preliminarily evaluate the effectiveness of this approach, we perform an evaluation with the ISPD2015 benchmarks. Table \ref{tab:regime} depicts this evaluation.

\begin{table}[h]
\caption{Evaluation of the proposed design classification approach over the ISPD2015 benchmarks.}
\label{tab:regime}
\centering
\scriptsize
\setlength{\tabcolsep}{4pt}
\begin{tabular}{cccccc}
Design & $S_p$ & $R_m$ & Class & Train $\Delta$\% & Eval $\Delta$\% \\
\midrule
\texttt{mgc\_fft\_1}  & $1.14{\times}10^{-3}$ & 9.27 & A & $\mathbf{0.53}$ & $0.20$ \\
\texttt{mgc\_pci\_bridge32\_a} & $1.96{\times}10^{-3}$ & 0.97 & B & $\mathbf{5.85}$ & $0.43$ \\
\texttt{mgc\_matrix\_mult\_1}     & $1.41{\times}10^{-4}$ & 4.77 & C & $\mathbf{0.04}$ & $0.03$ \\
\bottomrule
\end{tabular}
\smallskip
\end{table}


Each design lands in a different class, matching its outcome. \texttt{mgc\_fft\_1}
(A) has a high score and a clean reward, so full-graph SAGE is best and forcing a
subregion collapses its policy. \texttt{mgc\_pci\_bridge32\_a} (B) has a high
score but a noisy reward, full-graph training fails, while subregion training
cleans the signal and unlocks the largest gain. \texttt{mgc\_matrix\_mult\_1} (C)
has a low score from limited coverage and embedding collapse, so even subregion
training improves it only marginally.

\section{Conclusion}

We presented a design space exploration of GNN-PPO for post-legalization
placement refinement over 18 configurations across regular and irregular
ISPD2015 netlists. 
Regular designs expose three compounding pitfalls a credit-assignment bottleneck,
GNN over-smoothing, and tightly placed cells making subregion training most beneficial
on irregular,
noise-limited designs and only marginally on the regular
\texttt{mgc\_matrix\_mult\_1}, confirming that a netlist's regularity, reflected
in its cell spacing, indirectly indicates how much refinement it can still gain.
Guided by this, we proposed a lightweight classification based on the partition
score $S_p$ and the reward signal-to-noise ratio $R_m$ that selects full-graph or
subregion training from the netlist alone, calibrated on three designs and
pending validation on a broader suite.



\bibliographystyle{IEEEtran}
\bibliography{refs}

\end{document}